\documentclass[letterpaper, 10 pt, conference]{ieeeconf}
\usepackage{hyperref}
\IEEEoverridecommandlockouts

\usepackage{multirow}

\title{\LARGE \bf
Range and Bird's Eye View Fused Cross-Modal Visual Place Recognition
}

\author{Jianyi Peng, Fan Lu, Bin Li, Yuan Huang, Sanqing Qu, Guang Chen$^{*}$
\thanks{* corresponding author}
\thanks{Jianyi Peng, Fan Lu, Bin Li, Sanqing Qu and Guang Chen are with Tongji University, Shanghai 201804, China. E-mail: 2233589@tongji.edu.cn, lufan@tongji.edu.cn, libin2021@tongji.edu.cn, sanqingqu@tongji.edu.cn, guangchen@tongji.edu.cn.}
\thanks{Yuan Huang is with Automation, Beijing Institute of Control Engineering. E-mail: hy0103053428@gmail.com.}
\thanks{This work was supported by the National Key Research and Development Program of China (No. 2024YFE0211000), in part by the National Natural Science Foundation of China (No. 62372329), in part by Shanghai Scientific Innovation Foundation (No. 23DZ1203400), in part by Tongji-Qomolo Autonomous Driving Commercial Vehicle Joint Lab Project, and in part by Xiaomi Young Talents Program.}
}

\usepackage{graphicx}
\usepackage[linesnumbered,ruled,vlined]{algorithm2e}
\usepackage{amsmath}
\usepackage{stfloats}
\usepackage{amssymb}
\usepackage[dvipsnames]{xcolor}

\begin{document}

\maketitle
\thispagestyle{empty}
\pagestyle{empty}

\begin{abstract}
Image-to-point cloud cross-modal Visual Place Recognition (VPR) is a challenging task where the query is an RGB image, and the database samples are LiDAR point clouds. Compared to single-modal VPR, this approach benefits from the widespread availability of RGB cameras and the robustness of point clouds in providing accurate spatial geometry and distance information. However, current methods rely on intermediate modalities that capture either the vertical or horizontal field of view, limiting their ability to fully exploit the complementary information from both sensors. 
In this work, we propose an innovative initial retrieval + re-rank method that effectively combines information from range (or RGB) images and Bird's Eye View (BEV) images. Our approach relies solely on a computationally efficient global descriptor similarity search process to achieve re-ranking. Additionally, we introduce a novel similarity label supervision technique to maximize the utility of limited training data. Specifically, we employ points average distance to approximate appearance similarity and incorporate an adaptive margin, based on similarity differences, into the vanilla triplet loss. 
Experimental results on the KITTI dataset demonstrate that our method significantly outperforms state-of-the-art approaches. Code is available at \href{https://github.com/cppcute-pm/RangeBEV}{\texttt{github.com/cppcute-pm/RangeBEV}}.
\end{abstract}
\section{Introduction}
Image-to-point cloud cross-modal Visual Place Recognition (cross-modal VPR) involves querying a LiDAR point cloud database using an RGB image captured by a camera. Cameras are cost-effective and widely deployed in vehicles, making them ideal for online queries, while LiDAR provides precise spatial geometry and distance information. Cross-modal VPR addresses challenges such as environmental variations (e.g., lighting, weather, and seasonal changes) and eliminates the need for simultaneous mapping in visual SLAM systems. However, its retrieval performance lags behind single-modal VPR, primarily due to the significant modality gap between RGB images and LiDAR point clouds. RGB images capture dense color and texture details, whereas LiDAR point clouds provide accurate spatial data. Existing methods bridge this gap using intermediate modalities or similarity labels for supervision.

\begin{figure}[]
    \centering
    \includegraphics[scale=0.1]{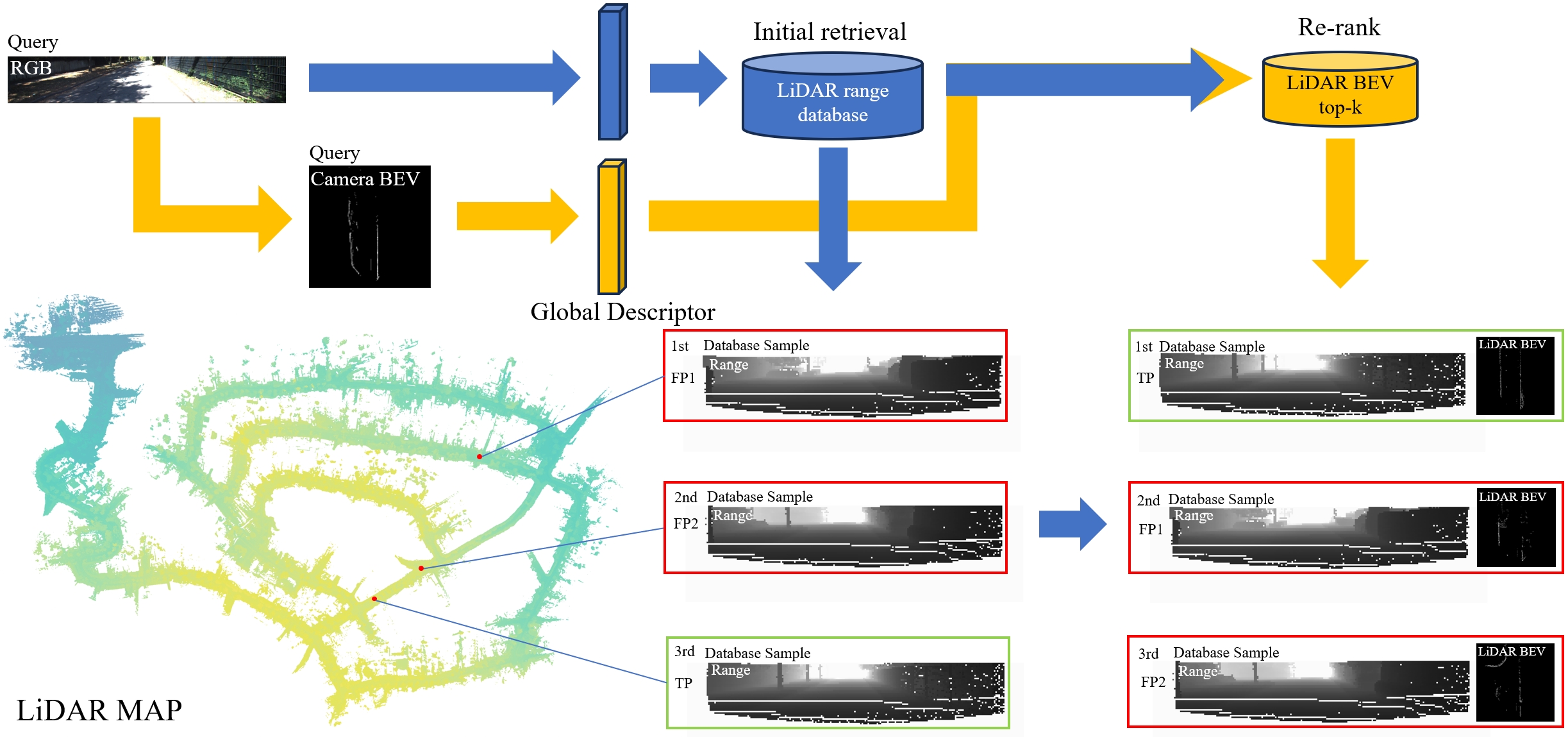}
    \caption{Illustration of our image-to-point cloud cross-modal visual place recognition. It's mainly composed of two separate similarity search process by only using global descriptors, in this way, we can effectively combine the information from range (or RGB) images and Bird's Eye View (BEV) images, significantly reducing the modality gap.}
    \label{fig:overview}
\end{figure}

Previous works have explored intermediate modalities to reduce the modality gap. For example, i3dLoc \cite{yin2021i3dloc} predicts pseudo range images from RGB images, while (LC)\textsuperscript{2} \cite{lee20232} generates depth maps from RGB images, which I2P-Rec \cite{zheng2023i2p} converts into pseudo point clouds for BEV images. ModaLink \cite{xie2024modalink} densifies depth projections from LiDAR point clouds to replace range images. Recent studies, such as LIP-Loc \cite{shubodh2024lip} and CMVM \cite{yao2024monocular}, align global descriptors of RGB and range images to reduce retrieval difficulty. Despite these efforts, the modality gap persists. To address this, we propose a strategy that combines the strengths of range (or RGB) images and BEV images, minimizing the gap's adverse effects.

Supervision with precise labels is another mainstream approach. (LC)\textsuperscript{2} \cite{lee20232} uses a sector area-based overlap ratio as a similarity label, combined with generalized contrastive loss. CMVM \cite{yao2024monocular} adopts a pixel-based overlap ratio from OverlapNet \cite{chen2021overlapnet} and predefined thresholds for binary labels. However, these methods suffer from inaccurate similarity approximations and lack continuous similarity label loss functions suitable for small datasets. To overcome these limitations, we propose a novel similarity label supervision method that provides more precise labels and ensures effective training with limited data.

In this work, we propose a fused pipeline leveraging range images and BEV images to enhance cross-modal VPR performance. As shown in Fig. \ref{fig:overview}, our method operates in two phases. First, we retrieve top-k candidates by computing global descriptor similarity between the query RGB image and range images of database submaps. Then, for the top-k candidates, we compute similarity between the camera BEV image and LiDAR BEV images. Finally, we perform a weighted summarization of the two ranking results. Automotive sensors are typically mounted horizontally, so range and RGB images capture vertical content, while BEV images provide a top-down view of spatial distribution. Our pipeline combines these modalities effectively, reducing computational costs.

Additionally, we propose a novel similarity label supervision method, including points average distance, a distance-based similarity generation method for more accurate appearance similarity approximation. We also adapt vanilla triplet loss for limited data training, ensuring robust supervision and improved performance.

Our contributions are as follows:
\begin{itemize}
  \item We propose an initial retrieval + re-rank pipeline that combines the strengths of range (or RGB) images and BEV images, using only global descriptors for robust VPR.
  \item We introduce a novel similarity label supervision approach, incorporating points average distance and generalized triplet loss, designed to extract meaningful patterns from limited datasets.
  \item Extensive experiments on the KITTI dataset demonstrate that our method outperforms state-of-the-art (SOTA) methods, setting a new benchmark in cross-modal VPR.
\end{itemize}
\section{Related Work}
In GNSS-based global positioning, vehicles rely on satellites and base stations for initial localization. However, satellite signals are often obstructed in urban areas, and base station coverage is limited in remote regions, increasing latency and hindering real-time applications. Visual Place Recognition (VPR) addresses these limitations by using visual sensors for robust global positioning. This section reviews single-modal and cross-modal VPR techniques, focusing on RGB images and point clouds, and discusses advancements in similarity label supervision.

\subsection{Single-Modal Visual Place Recognition}
Single-modal VPR relies on a single sensor type for place recognition. Early methods used hand-crafted features like VLAD \cite{jegou2010aggregating} and SIFT \cite{lowe1999object} for RGB image-based VPR, leveraging cameras' cost-effectiveness and widespread use. These methods assume similar local features share appearances, with feature centers representing general scene cues. Residuals between local features and centers are aggregated into global descriptors. With deep learning, CNNs emerged for feature extraction. NetVLAD \cite{arandjelovic2016netvlad}, a neural adaptation of VLAD, refined feature centers and introduced soft weight assignment, improving performance. Generalized Mean Pooling (GeM) \cite{radenovic2018fine} further enhanced feature aggregation. For point cloud-based VPR, PointNetVLAD \cite{uy2018pointnetvlad} combined PointNet \cite{qi2017pointnet} with VLAD, enabling effective LiDAR data use. LiDAR provides precise distance measurements, avoiding RGB-based issues like perception aliasing and weather sensitivity. Advanced backbones, such as Transformer \cite{waswani2017attention}, improved feature extraction, while aggregators like (\cite{izquierdo2024optimal}, \cite{lu2024supervlad}, \cite{khaliq2024vlad}, \cite{ali2024boq}), enhanced global descriptor discrimination. Large-scale datasets like (\cite{warburg2020mapillary}, \cite{berton2022rethinking}, \cite{ali2022gsv}, \cite{alibeigi2023zenseact}), have advanced practical VPR applications, with frameworks like (\cite{berton2025meshvpr}, \cite{berton2024earthloc}, \cite{kolmet2022text2pos}, \cite{keetha2023anyloc}, \cite{vivanco2024geoclip}, \cite{xu2024addressclip}), expanding its scope.

\subsection{Cross-Modal Visual Place Recognition}
Cross-modal VPR involves querying a LiDAR point cloud database using an RGB image's global descriptor to retrieve the most similar location. The primary challenge is bridging the modality gap: RGB images capture color and texture, while point clouds provide spatial and distance data. Early approaches, such as PlainEBD \cite{cattaneo2020global}, aligned global descriptors directly using metric learning and knowledge distillation. Subsequent works, like i3dLoc \cite{yin2021i3dloc}, introduced range images as an intermediate modality to improve retrieval. (LC)\textsuperscript{2} \cite{lee20232} removed color information by estimating depth from RGB images, while ModaLink \cite{xie2024modalink} proposed field-of-view (FoV) clipping to enhance content overlap, though this struggles with viewpoint variations. Recent studies, such as CMVM \cite{yao2024monocular}, split 360° LiDAR point clouds into sections, improving retrieval accuracy despite increased database size. I2P-Rec \cite{zheng2023i2p} utilized BEV images, generated via depth estimation, achieving competitive performance. While range images excel in vertical geometric detail, BEV images provide a comprehensive horizontal perspective but lack fine-grained detail. This work integrates both modalities for a more robust solution.

\subsection{Similarity Label Supervision}
Similarity label supervision enhances retrieval in metric learning. Traditional methods rely on binary labels derived from GPS or UTM distances, often overlooking visual content overlap. The MSLS \cite{warburg2020mapillary} dataset improved supervision by incorporating heading angles. BEV\textsuperscript{2}PR \cite{ge2024bev2pr} adjusted GPS coordinates along the heading direction but still used binary labels. GCL \cite{leyva2023data} introduced refined labeling using sector area overlap for outdoor scenes and point cloud mutual nearest neighbor overlap for indoor scenes, along with a modified contrastive loss for continuous labels. OverlapNet \cite{chen2021overlapnet} used range image overlap for supervision. Our work proposes an enhanced similarity labeling approach to improve model performance.
\begin{figure*}[t]
    \centering
    \includegraphics[scale=0.14]{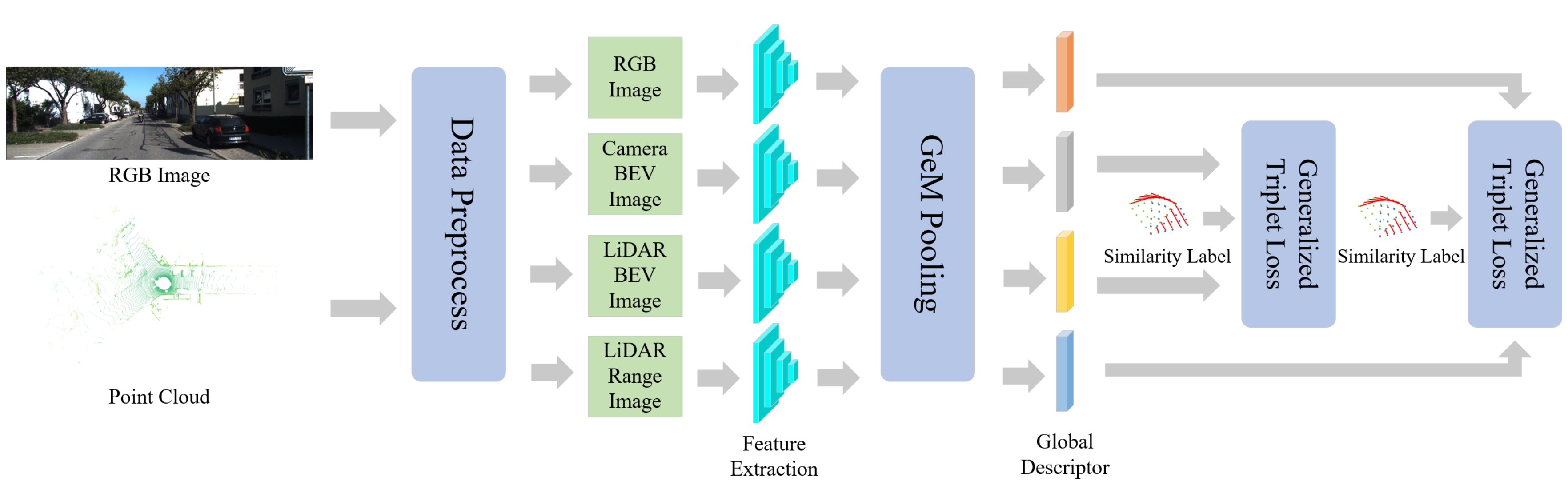}
    \caption{The training pipeline to produce the range (or RGB) and BEV descriptors. raw data (e.g., LiDAR point cloud, camera RGB image) are preprocessed to reduce modality differences and improve the overlap in visual content. Featrue maps are generated by extracting features from RGB, LiDAR range, camera BEV and LiDAR BEV, which are latter aggregated by the Generalized Mean (GeM) pooling to abtain global descriptors. It's worth noting that we use the points average distance together with a generalized triplet loss to supervise the learning process and fully utilize the limited training data}
    \label{fig:pipeline}
\end{figure*}
\section{Methods}
\subsection{Overview}
This research presents a framework, seen in Fig. \ref{fig:overview}, that utilizes a two-stage approach consisting of initial retrieval and subsequent re-ranking to improve the precision of location recognition. Significantly, global descriptors are solely employed in both phases. The training pipeline, seen in Fig. \ref{fig:pipeline}, commences with data preparation to produce four image types: RGB images, range images obtained from LiDAR point clouds, BEV images created from RGB photos, and BEV images generated from LiDAR point clouds. These images are processed through ResNet50 \cite{he2016deep} backbones to extract feature maps, which are then aggregated by using the GeM pooling \cite{radenovic2018fine} method to obtain four kinds of global descriptors. A unique similarity label generation method and a modified triplet loss \cite{schroff2015facenet} are utilized to enhance contrastive learning by supervising the relationship between RGB images and LiDAR range images, as well as between camera BEV images and LiDAR BEV images. The specifics of each component will be further upon in later sections.

\subsection{Feature Extraction and Aggregation}
Utilizing the data preprocessing methods described in LIP-Loc \cite{shubodh2024lip} and I2P-Rec \cite{shubodh2024lip}, we produce the subsequent inputs: $I_{RGB} \in \mathbb{R}^{H_1{\times}W_1{\times}3}$, $I_{range} \in \mathbb{R}^{H_2{\times}W_2{\times}3}$, $I_{camera\_BEV} \in \mathbb{R}^{H_3{\times}W_3{\times}3}$, $I_{LiDAR\_BEV} \in \mathbb{R}^{H_3{\times}W_3{\times}3}$. It's worth noting that we filter out ambiguous depth information and remove the ground points during data preprocessing to reduce noise, the details are in the Appendix. Subsequently, the ResNet50 encoder network processes these inputs to extract feature maps, resulting in the outputs: $F_{RGB} \in \mathbb{R}^{7{\times}7{\times}C_1}$, $F_{range} \in \mathbb{R}^{4{\times}24{\times}C_2}$, $F_{camera\_BEV} \in \mathbb{R}^{4{\times}4{\times}C_3}$, $F_{LiDAR\_BEV} \in \mathbb{R}^{4{\times}4{\times}C_3}$. Subsequently, a multi-layer perceptron (MLP) is utilized as a dimensionality reduction network to map all feature mappings into a cohesive 256-dimensional space. The generalized mean pooling (GeM pooling) approach is ultimately employed to consolidate these feature maps, yielding the global descriptors: $G_{RGB} \in \mathbb{R}^{256}$, $G_{range} \in \mathbb{R}^{256}$, $G_{camera\_BEV} \in \mathbb{R}^{256}$, $G_{LiDAR\_BEV} \in \mathbb{R}^{256}$.

\subsection{Similarity Label Supervision}
This section introduces our similarity label supervision approach, which consists of two key components: similarity label creation and the similarity loss function.

GCL \cite{leyva2023data} estimates visual similarity using the ratio of overlapping sector areas. However, this method has limitations. First, calculating overlapping areas is computationally complex, preventing the use of efficient similarity search engines like Faiss \cite{douze2024faiss} and requiring an overlap ratio matrix with $O(n^2)$ spatial complexity, which is impractical for large-scale VPR datasets. Second, the sector area overlap ratio is a coarse approximation of appearance similarity, as it fails to capture the intricate distribution of visual features in 3D space.
 
\begin{figure}[]
    \centering
    \includegraphics[scale=0.24]{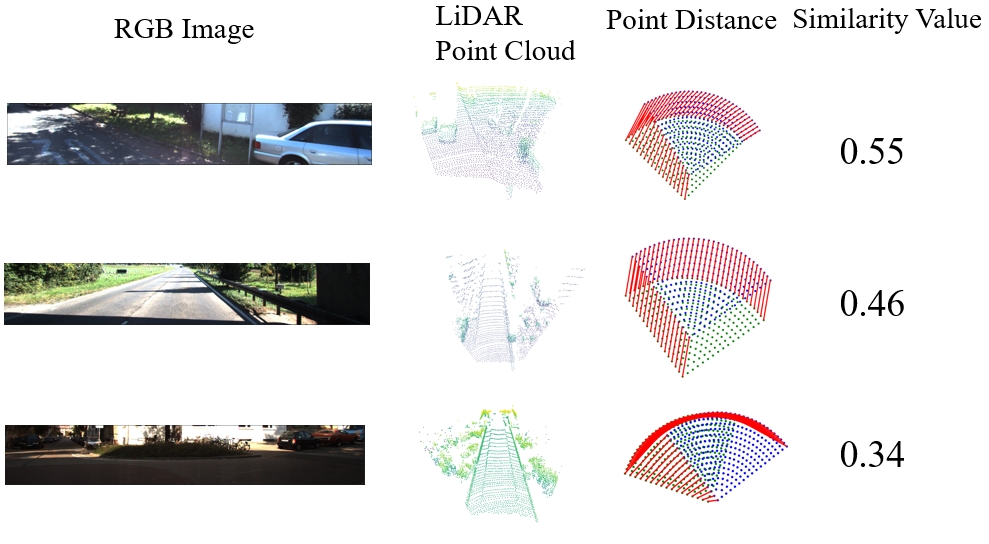}
    \caption{Examples of three submaps with similarity value in a decreasing order. RGB image and LiDAR Point Cloud is cropped to maximize the visual content overlapping. In the third column, red lines connecting corresponding blue and green points indicate the points distances, which are later processed for obtaining the final similarity value.}
    \label{fig:SimLabelSupVis}
\end{figure}
For similarity label creation, we propose a distance-based similarity label generation method called points average distance, which provides a more precise similarity measure. As illustrated in Fig. \ref{fig:SimLabelSupVis}, we replace a single point coordinate with a set of uniformly sampled points within a defined area. Each point's location is fixed relative to the ego vehicle's coordinate system, and points in different sets have a one-to-one correspondence. The similarity value is computed as follows:
\begin{equation}
\begin{gathered}
D_{avg}(i,j)={\frac{1}{n}}{\sum_{k=1}^{n}}{{||{Coord_{i,k}-Coord_{j,k}}||}_2} \\
{Sim(i,j)=
\begin{cases}
\frac{D_{th}-D_{avg}(i,j)}{D_{th}},& if D_{avg}(i,j)<{D_{th}} \\
0,& else
\end{cases}}
\end{gathered}
\end{equation}
\indent Here, $D_{avg}(i,j)$ is the average Euclidean distance between corresponding point pairs, and $D_{th}$ is a threshold determining the final similarity value $Sim(i,j)$.

For similarity loss function, we propose generalized triplet loss, which adapts the margin based on similarity labels. Unlike conventional triplet loss \cite{schroff2015facenet} with a fixed margin, our approach uses an adaptive margin, accommodating the non-binary nature of similarity labels.

To elaborate, we consider three samples: an anchor sample $x_a$, a random sample $x_1$, and another random sample $x_2$. The similarity between $x_a$ and $x_1$ is denoted as $sim_{a1}$, while the similarity between $x_a$ and $x_2$ is denoted as $sim_{a2}$. If $sim_{a1} > sim_{a2}$, $x_1$ is treated as a relative positive sample $x_{rp}$, and $x_2$ is treated as a relative negative sample $x_{rn}$. The difference in feature distances between these samples is adjusted based on their similarity variation. We introduce a similarity difference term, combined with a base margin ${\alpha}_{base}$, to define the adaptive margin. The final loss function is formulated as:
\begin{equation}
\mathcal{L}=max(D(a, rp)-D(a, rn)+{\alpha}_{base}\cdot{(sim_{arp}-sim_{arn})},0)
\end{equation}

\subsection{Initial Retrieval and Re-rank}
Aligning the global descriptors of range images and RGB images guarantees uniformity in visual information along the vertical axis. Nonetheless, this alignment alone is inadequate for the model to deduce the horizontal spatial distribution and conduct further comparisons. To mitigate this constraint, we integrate BEV images from camera and LiDAR sensors. Inspired by the prevalent initial retrieval and re-ranking pipeline \cite{shao2023global}, we present a computationally efficient solution that eliminates the need for an auxiliary feature matching network or spatial verification process.
\begin{figure}[]
    \centering
    \includegraphics[scale=0.11]{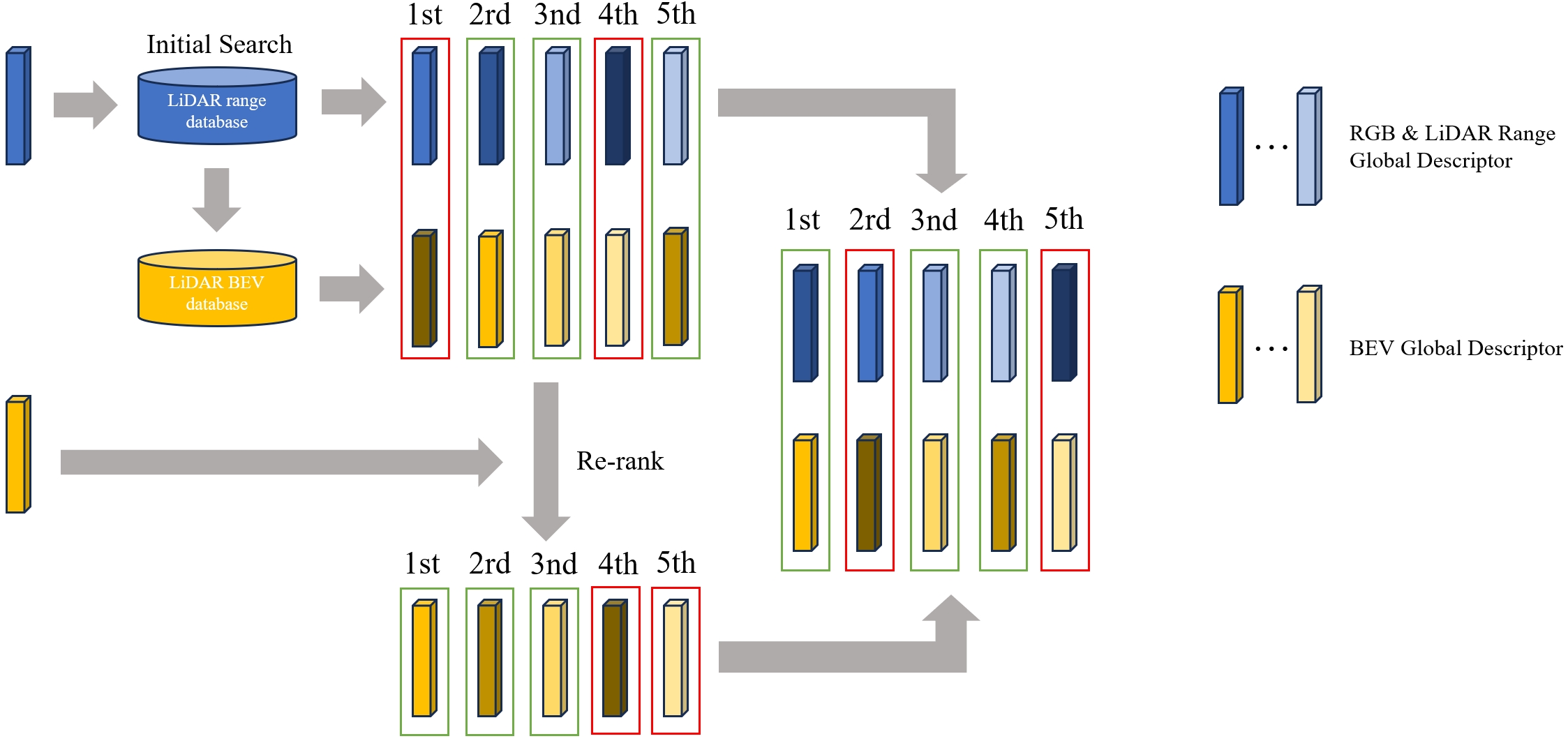}
    \caption{The initial retrieval + re-rank pipeline. In the two-phase similarity search, global descriptors with higher similarity, indicated by closer similarity in color, are ranked higher. By combining the rankings from both phases, we improve the precision of retrieval. This results in true positive samples being ranked higher (indicated by global descriptors in green boxes) and false positive samples being ranked lower (indicated by global descriptors in red boxes).}
    \label{fig:rerank}
\end{figure}
As depicted in Fig. \ref{fig:rerank}, our methodology is executed as follows: Initially, we utilize the global descriptor $G_{RGB}$ to do retrieval within the $G_{range}$ database, yielding the top-k most analogous LiDAR submaps as potential matches, with the highest rank being documented. In our trials, the variable $k$ is established at 60. Subsequently, we employ the global descriptor $G_{camera\_BEV}$ as a query to search among the top-k candidates from the $G_{LiDAR\_BEV}$ collection, resulting in a secondary ranking. The two ranks are ultimately weighted and aggregated to yield the final re-ranked outcome.

Our methodology circumvents the computationally intensive tasks of feature matching (\cite{sarlin2020superglue}, \cite{sun2021loftr}, \cite{lindenberger2023lightglue}, \cite{wang2024efficient}) and spatial verification (\cite{lee2022correlation}, \cite{xue2022efficient}) found in conventional re-ranking algorithms (\cite{wang2022transvpr}, \cite{zhang2023etr}, \cite{zhu2023r2former}, \cite{hausler2021patch}) in VPR. Moreover, unlike query expansion \cite{radenovic2018fine} approaches typically employed in image retrieval, our approach necessitates searching solely inside the top-k candidates instead of the full database. Experimental findings indicate that combining global descriptors from adjacent samples diminishes retrieval performance. This is due to the distinctive feature distribution that maintains the same neighborhood relationships \cite{luo2023bevplace} as the geographic distribution of scenes, easily introducing noise when there exists a false positive sample in the neighborhood \cite{zaffar2024estimation}.
\section{Experiments}

\begin{table*}[!ht]
    \centering
    \caption{RECALL@N (\%) ON THE KITTI DATASET. THE BEST RESULTS IN EACH COLUMN ARE HIGHLIGHTED IN BOLD, AND THE SECONDS ARE UNDERLINED.}
    \label{tab:CompSOTA}
    \resizebox{\textwidth}{!}{
    \begin{tabular}{c|c|c|c|c|c|c}
    \hline
         & KITTI-00 & KITTI-02 & KITTI-05 & KITTI-06 & KITTI-07 & KITTI-08 \\ \hline
        Method & R@1 / R@5 / R@1\% & R@1 / R@5 / R@1\% & R@1 / R@5 / R@1\% & R@1 / R@5 / R@1\% & R@1 / R@5 / R@1\% & R@1 / R@5 / R@1\% \\ \hline
        PlainEBD \cite{cattaneo2020global} & 19.93 / 31.71 / 69.06 & 16.50 / 27.48 / 55.93    & 30.64 / 46.11 / 71.10 & 25.34 / 35.88 / 47.68 & 39.95 / 59.30 / 70.93  & 20.49 / 35.96 / 72.22 \\
        (LC)\textsuperscript{2} \cite{lee20232} & 30.65 / 47.76 / 84.52 & 23.06 / 35.72 / 67.41 & 40.49 / 54.58 / 80.62 & 39.78 / 58.22 / 71.48 & 52.50 / 68.03 / 84.01 & 38.42 / 54.07 / 87.30 \\
        I2P-Rec \cite{zheng2023i2p} & 44.84 / 59.55 / 90.09 & 28.00 / 42.33 / 71.68 & 48.32 / 64.61 / 88.52 & 43.51 / 63.12 / 72.30 & 63.03 / 73.39 / 78.93 & 45.91 / 62.44 / 91.65 \\
        VXP \cite{li2024vxp} & 24.22 / 38.16 / 80.00 & 17.72 / 30.83 / 64.64 & 32.81 / 50.63 / 78.16 & 29.97 / 41.96 / 52.50 & 43.69 / 61.31 / 76.29 & 24.01 / 37.68 / 79.17 \\
        ModaLink \cite{xie2024modalink} & 90.31 / 95.53 / 99.58 & 70.18 / 81.85 / 96.74 & 90.00 / 95.00 / \underline{99.31} & 86.19 / 96.64 / \underline{100.0} & 96.55 / 99.27 / \underline{100.0} & 86.97 / \underline{95.75} / \underline{99.93} \\
        CMVM \cite{yao2024monocular} & \underline{93.13} / \underline{96.83} / \underline{99.74} & \underline{76.98} / \underline{86.44} / \underline{97.51} & \underline{90.76} / \underline{96.09} / 99.24 & \underline{89.28} / \underline{97.73} / 99.18 & \underline{98.09} / \underline{99.46} / 99.82 & \underline{90.64} / 95.23 / 99.63 \\
        Ours & \textbf{99.03} / \textbf{99.91} / \textbf{100.0} & \textbf{95.07} / \textbf{98.22} / \textbf{99.66} & \textbf{98.91} / \textbf{99.53} / \textbf{100.0} & \textbf{97.00} / \textbf{100.0} / \textbf{100.0} & \textbf{99.91} / \textbf{100.0} / \textbf{100.0} & \textbf{98.21} / \textbf{99.51} / \textbf{99.93} \\ \hline
    \end{tabular}}
\end{table*}

In this section, we first provide the implementation details, followed by a description of the dataset and evaluation metrics. We then compare the retrieval performance of our method with baseline approaches, and conclude with an ablation study to validate the effectiveness of the proposed methods.

\subsection{Experimental Settings}

\textbf{Implementation Details}. Experiments were conducted on an Intel i9-10920X CPU and an NVIDIA RTX 3090 GPU. We use the Depth Anything model \cite{yang2024depth}, pre-trained on the KITTI dataset, for monocular depth prediction. As in \cite{leyva2023data}, the sector area is utilized for similarity label generation, with a circumcentric angle of 90° (matching the camera’s horizontal field of view) and a radius of 10m, which corresponds to the true positive distance threshold in prior cross-modal VPR work. The point set distance threshold is set to 7.5m, and the base margin is set to 0.6. It is important to note that we use pretrained ResNet50 weights on the Cityscapes dataset to extract features from RGB images.  

\textbf{Dataset}. The KITTI Odometry dataset \cite{geiger2012we}, a widely used dataset in cross-modal Visual Place Recognition tasks, serves as the primary dataset for training and evaluation. The dataset consists of 22 sequences, where the pose for each RGB image frame and LiDAR point cloud frame is annotated using combined location information from INS and GPS. Following the setup in \cite{yao2024monocular}, we use the last 11 sequences for training and the first 11 sequences for testing.

\textbf{Evaluation Protocol}. For evaluation, we employ the \textbf{recall at Top-N} and \textbf{recall at Top-1\%} as our metrics. Following previous works \cite{zheng2023i2p}, \cite{xie2024modalink}, and \cite{yao2024monocular}, a match is considered successful if the distance between the query’s ground truth location and the candidate’s ground truth location is within a true positive distance threshold, \( t = 10 \text{m} \).

\subsection{Comparison with State-of-the-Arts Methods}
In this section, we compare our method with several SOTA image-to-point cloud place recognition methods: PlainEBD \cite{cattaneo2020global}, I2P-Rec \cite{zheng2023i2p}, VXP \cite{li2024vxp}, ModaLink \cite{xie2024modalink} and CMVM \cite{yao2024monocular}. Tabel \ref{tab:CompSOTA} shows the quantitative results on KITTI Odometry dataset. We can see that ModaLink and CMVM both show very high recognition performance compared with other baseline methods and they both use the RGB Image and the LiDAR Range Image as the intermediate modality input, same as our original setting. However, with the aid of the proposed similarity label supervision and BEV global descriptor re-ranking, our method achieve an impressive State-of-the-Arts performance in every sequence. We surpass the previous SOTA by 18.09\% on Recall@1 in KITTI-02 and reach 99\% on Recall@1 in KITTI-00 and KITTI-07. The proposed similarity label supervision can approximate the appearance similarity very well and successifully make the model identify the appearance nuances among samples by introducing the generalized triplet loss.

\subsection{Ablation Study}
In this section, we evaluate the effectiveness of the proposed components in our method through several experiments.

\begin{table}[!t]
    \centering
    \caption{ABLATION STUDIES ON KITTI-00 FOR SIMILARITY LABEL SUPERVISION AND BEV DESCRIPTOR RE-RANK}
    \label{tab:DiffComp}
    \begin{tabular}{c|c|c|c}
    \hline
         Vallina & Sim SV & BEV Re-rank & R@1 / R@5 / R@1\% \\ \hline
         \checkmark &  &  & 68.40 / 82.18 / 97.01 \\
         \checkmark & \checkmark &  & 98.63 / 99.82 / 100.0 \\
         \checkmark &  & \checkmark & 81.83 / 90.73 / 97.60 \\
         \checkmark & \checkmark & \checkmark & 99.03 / 99.91 / 100.0 \\ \hline
    \end{tabular}
\end{table}

\textbf{Ablation on Different Components}. We conducted experiments by removing their individual components, including similarity label supervision (Sim SV) and BEV global descriptor re-ranking (BEV Re-rank). As shown in Table \ref{tab:DiffComp}, introducing similarity label supervision leads to a significant performance improvement of 30.23\% in Recall@1 on the KITTI-00 sequence. This demonstrates the importance of the points average distance and generalized triplet loss in the framework. Furthermore, using BEV global descriptors from both the RGB and LiDAR sensors for a second retrieval within the top-k candidates leads to a performance increase of 13.43\% in Recall@1. Combining these two methods achieves a remarkable Recall@1 of 99.03\%, validating the overall effectiveness of our pipeline.

\textbf{Ablation on Similarity Label Supervision}. 
In this part, we separately evaluate the Similarity Label Generation and Similarity Loss Function.

For similarity label generation, We compare the points average distance with other similarity label generation methods in Table \ref{tab:DiffSLG}. (position, velocity vector) \cite{ge2024bev2pr} method calculates binary labels based on the distance between points' locations and heading angles, leading to poor performance. Point cloud mnn \cite{leyva2023data}, designed for indoor point cloud submaps, doesn't perform well for outdoor scenes. Exponential negative distance, a method we proposed earlier, uses an exponentiation operation on the distance between two ground truth locations but introduces an additional hyperparameter. This method yields performance comparable to the area overlap \cite{leyva2023data} method. By introducing the average distance of point sets, our points average distance method consistently outperforms state-of-the-art approaches in both KITTI-00 and KITTI-02. Appendix analyzes the impact of the distance threshold.

\begin{table}[!t]
    \centering
    \caption{ABLATION STUDIES ON DIFFERENT SIMILARITY LABEL GENERATION METHODS}
    \label{tab:DiffSLG}
    \resizebox{8.5cm}{!}{
    \begin{tabular}{c|c|c}
    \hline
        & KITTI-00 & KITTI-02 \\ \hline
        Method & R@1 / R@5 / R@1\% & R@1 / R@5 / R@1\% \\ \hline
        (Position, Velocity Vector) \cite{ge2024bev2pr} & 77.45 / 86.17 / 96.32 & 61.94 / 73.12 / 87.99 \\
        Area Overlap \cite{leyva2023data} & 98.26 / 99.80 / 100.0 & 94.23 / 97.55 / 99.55 \\
        Point Cloud NN \cite{leyva2023data} & 95.75 / 98.57 / 99.91 & 86.35 / 91.44 / 95.67 \\
        Exponential Negative Distance & 98.00 / 99.87 / 100.0 & 94.66 / 98.41 / 99.49 \\
        Points Average Distance & 99.03 / 99.91 / 100.0 & 95.07 / 98.22 / 99.66 \\ \hline
    \end{tabular}
    }
\end{table}

\begin{table}[!t]
    \centering
    \caption{ABLATION STUDIES ON DIFFERENT LOSS FUNCTIONS}
    \label{tab:DiffLoss}
    \resizebox{8.5cm}{!}{
    \begin{tabular}{c|c|c}
    \hline
        & KITTI-00 & KITTI-02 \\ \hline
        Method & R@1 / R@5 / R@1\% & R@1 / R@5 / R@1\% \\ \hline
        Generalized Contrastive Loss \cite{leyva2023data} & 83.46 / 93.50 / 98.72 & 64.60 / 77.56 / 88.39 \\
        Triplet Margin Loss \cite{schroff2015facenet} & 96.45 / 98.79 / 99.91 & 85.71 / 91.29 / 97.04  \\
        Generalized Triplet Margin loss & 99.03 / 99.91 / 100.0 & 95.07 / 98.22 / 99.66 \\ \hline
    \end{tabular}
    }
\end{table}

\begin{table}[!t]
    \centering
    \caption{ABLATION STUDIES ON FUSION METHODS}
    \label{tab:AblationFM}
    \resizebox{8.5cm}{!}{
    \begin{tabular}{c|c|c}
    \hline
        & KITTI-00 & KITTI-02 \\ \hline
        Method & R@1 / R@5 / R@1\% & R@1 / R@5 / R@1\% \\ \hline
        Cat & 98.85 / 99.98 / 100.0 & 94.77 / 97.79 / 99.94 \\
        Global Descriptor + Attention \cite{zhang2024mvse} & 97.36 / 99.63 / 100.0 & 89.62 / 95.56 / 99.72 \\
        Local Feature + Attention & 98.48 / 99.76 / 100.0 & 92.83 / 96.82 / 100.0 \\
        Global Descriptor + Re-rank & 99.03 / 99.91 / 100.0 & 95.07 / 98.22 / 99.66 \\ \hline
    \end{tabular}
    }
\end{table}

For similarity loss function, in Table \ref{tab:DiffLoss}, we compare generalized triplet loss with vanilla triplet loss and generalized contrastive loss. The results show that even vanilla triplet loss significantly outperforms generalized contrastive loss, as the latter cannot effectively capture the relative relationships between anchor-positive and anchor-negative pairs, making it unsuitable for training on smaller datasets with adjacent feature distributions. Furthermore, the introduction of adaptive margin results in a 9.36\% increase in Recall@1 for KITTI-02, demonstrating the importance of fine-grained feature similarity, additional ablation on the base margin is in the Appendix.

\begin{figure}[!t]
    \centering
    \includegraphics[scale=0.18]{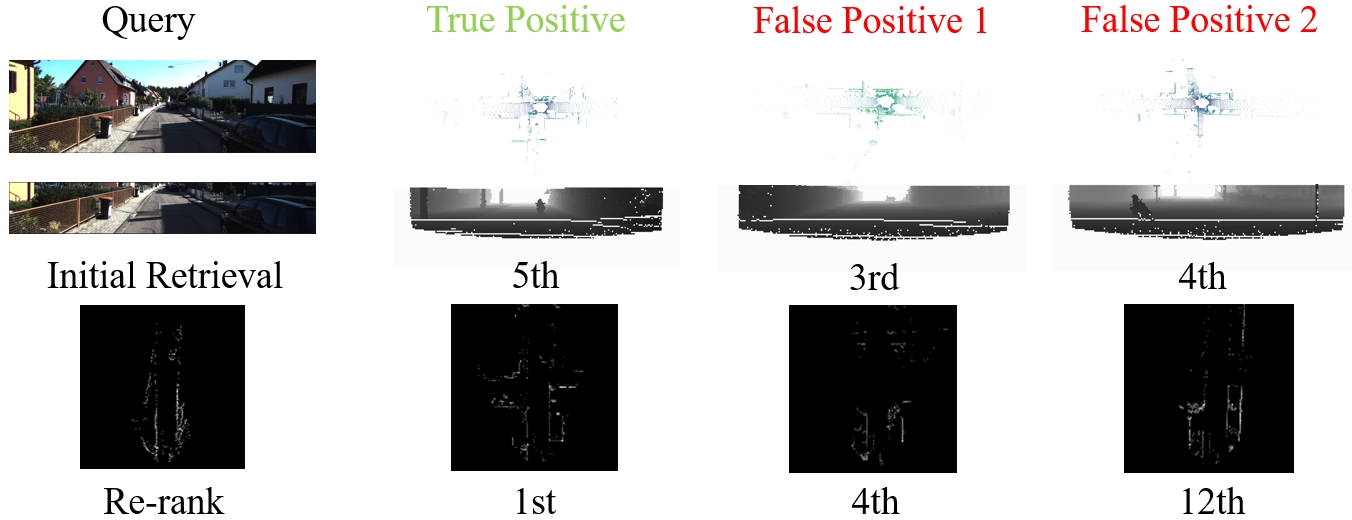}
    \caption{Examples of initial retrieval and re-rank results on KITTI dataset.}
    \label{fig:RerankVis}
\end{figure}

\textbf{Ablation on Re-rank}. 
Aligning information from the horizontal and vertical directions is a non-trivial task. As shown in Table \ref{tab:AblationFM}, besides using BEV global descriptors for re-ranking, we also explore other fusion methods. One approach involves directly concatenating the two global descriptors for retrieval, followed by applying a transformer block to extract context between the two descriptors, as done in \cite{zhang2024mvse}. Another method involves using cross-attention between the RGB and BEV feature maps before pooling and concatenation. While concatenating the two global descriptors shows a slight performance improvement, the effect is minimal. This indicates that fusion-based approaches require more complex designs to effectively combine the two perspectives. As displayed in Fig. \ref{fig:RerankVis}, our BEV global descriptor re-rank method successfully avoids viewpoint interference and increase the rank of true positive by separately using vertical and horizontal visual information. Additionally, in the Appendix, we do some ablation experiments to emphasize the importance of the ground and noise removal process.

\section{Conclusion}
In this paper, we address the challenge of image-to-point cloud cross-modal VPR and propose an innovative initial retrieval + re-rank method. This method effectively combines information from range (or RGB) images and BEV images in a computationally efficient manner to bridge the modality gap. Our approach exclusively uses a global descriptor similarity search process for re-ranking, thereby avoiding mutual interference between the two modalities.
Furthermore, we present a novel similarity label supervision technique to optimize the use of limited training data. By introducing the points average distance metric, we closely approximate appearance similarity, and the generalized triplet loss dynamically adjusts the margin based on the similarity difference between sample pairs.
Experimental results on the KITTI dataset validate that our method achieves significant improvements over state-of-the-art approaches. Future research will explore a more practical VPR scenario where the query is an RGB image and the database samples consist of both LiDAR point clouds and RGB images.


\bibliographystyle{IEEEtran}
\bibliography{IEEEfull}

\begin{thebibliography}{10}
\providecommand{\url}[1]{#1}
\csname url@samestyle\endcsname
\providecommand{\newblock}{\relax}
\providecommand{\bibinfo}[2]{#2}
\providecommand{\BIBentrySTDinterwordspacing}{\spaceskip=0pt\relax}
\providecommand{\BIBentryALTinterwordstretchfactor}{4}
\providecommand{\BIBentryALTinterwordspacing}{\spaceskip=\fontdimen2\font plus
\BIBentryALTinterwordstretchfactor\fontdimen3\font minus \fontdimen4\font\relax}
\providecommand{\BIBforeignlanguage}[2]{{%
\expandafter\ifx\csname l@#1\endcsname\relax
\typeout{** WARNING: IEEEtran.bst: No hyphenation pattern has been}%
\typeout{** loaded for the language `#1'. Using the pattern for}%
\typeout{** the default language instead.}%
\else
\language=\csname l@#1\endcsname
\fi
#2}}
\providecommand{\BIBdecl}{\relax}
\BIBdecl

\bibitem{yin2021i3dloc}
P.~Yin, L.~Xu, J.~Zhang, H.~Choset, and S.~Scherer, ``i3dloc: Image-to-range cross-domain localization robust to inconsistent environmental conditions,'' \emph{arXiv preprint arXiv:2105.12883}, 2021.

\bibitem{lee20232}
A.~J. Lee, S.~Song, H.~Lim, W.~Lee, and H.~Myung, ``(lc)$^{2}$: Lidar-camera loop constraints for cross-modal place recognition,'' \emph{IEEE Robotics and Automation Letters}, vol.~8, no.~6, pp. 3589--3596, 2023.

\bibitem{zheng2023i2p}
S.~Zheng, Y.~Li, Z.~Yu, B.~Yu, S.-Y. Cao, M.~Wang, J.~Xu, R.~Ai, W.~Gu, L.~Luo \emph{et~al.}, ``I2p-rec: Recognizing images on large-scale point cloud maps through bird's eye view projections,'' in \emph{2023 IEEE/RSJ International Conference on Intelligent Robots and Systems (IROS)}.\hskip 1em plus 0.5em minus 0.4em\relax IEEE, 2023, pp. 1395--1400.

\bibitem{xie2024modalink}
W.~Xie, L.~Luo, N.~Ye, Y.~Ren, S.~Du, M.~Wang, J.~Xu, R.~Ai, W.~Gu, and X.~Chen, ``Modalink: Unifying modalities for efficient image-to-pointcloud place recognition,'' \emph{arXiv preprint arXiv:2403.18762}, 2024.

\bibitem{shubodh2024lip}
S.~Shubodh, M.~Omama, H.~Zaidi, U.~S. Parihar, and M.~Krishna, ``Lip-loc: Lidar image pretraining for cross-modal localization,'' in \emph{Proceedings of the IEEE/CVF Winter Conference on Applications of Computer Vision}, 2024, pp. 948--957.

\bibitem{yao2024monocular}
G.~Yao, X.~Li, L.~Fu, and Y.~Pan, ``Monocular visual place recognition in lidar maps via cross-modal state space model and multi-view matching,'' \emph{arXiv preprint arXiv:2410.06285}, 2024.

\bibitem{chen2021overlapnet}
X.~Chen, T.~L{\"a}be, A.~Milioto, T.~R{\"o}hling, O.~Vysotska, A.~Haag, J.~Behley, and C.~Stachniss, ``Overlapnet: Loop closing for lidar-based slam,'' \emph{arXiv preprint arXiv:2105.11344}, 2021.

\bibitem{jegou2010aggregating}
H.~J{\'e}gou, M.~Douze, C.~Schmid, and P.~P{\'e}rez, ``Aggregating local descriptors into a compact image representation,'' in \emph{2010 IEEE computer society conference on computer vision and pattern recognition}.\hskip 1em plus 0.5em minus 0.4em\relax IEEE, 2010, pp. 3304--3311.

\bibitem{lowe1999object}
D.~G. Lowe, ``Object recognition from local scale-invariant features,'' in \emph{Proceedings of the seventh IEEE international conference on computer vision}, vol.~2.\hskip 1em plus 0.5em minus 0.4em\relax Ieee, 1999, pp. 1150--1157.

\bibitem{arandjelovic2016netvlad}
R.~Arandjelovic, P.~Gronat, A.~Torii, T.~Pajdla, and J.~Sivic, ``Netvlad: Cnn architecture for weakly supervised place recognition,'' in \emph{Proceedings of the IEEE conference on computer vision and pattern recognition}, 2016, pp. 5297--5307.

\bibitem{radenovic2018fine}
F.~Radenovi{\'c}, G.~Tolias, and O.~Chum, ``Fine-tuning cnn image retrieval with no human annotation,'' \emph{IEEE transactions on pattern analysis and machine intelligence}, vol.~41, no.~7, pp. 1655--1668, 2018.

\bibitem{uy2018pointnetvlad}
M.~A. Uy and G.~H. Lee, ``Pointnetvlad: Deep point cloud based retrieval for large-scale place recognition,'' in \emph{Proceedings of the IEEE conference on computer vision and pattern recognition}, 2018, pp. 4470--4479.

\bibitem{qi2017pointnet}
C.~R. Qi, H.~Su, K.~Mo, and L.~J. Guibas, ``Pointnet: Deep learning on point sets for 3d classification and segmentation,'' in \emph{Proceedings of the IEEE conference on computer vision and pattern recognition}, 2017, pp. 652--660.

\bibitem{waswani2017attention}
A.~Waswani, N.~Shazeer, N.~Parmar, J.~Uszkoreit, L.~Jones, A.~Gomez, L.~Kaiser, and I.~Polosukhin, ``Attention is all you need,'' in \emph{NIPS}, 2017.

\bibitem{izquierdo2024optimal}
S.~Izquierdo and J.~Civera, ``Optimal transport aggregation for visual place recognition,'' in \emph{Proceedings of the IEEE/CVF Conference on Computer Vision and Pattern Recognition}, 2024, pp. 17\,658--17\,668.

\bibitem{lu2024supervlad}
\BIBentryALTinterwordspacing
F.~Lu, X.~Zhang, C.~Ye, S.~Dong, L.~Zhang, X.~Lan, and C.~Yuan, ``Super{VLAD}: Compact and robust image descriptors for visual place recognition,'' in \emph{The Thirty-eighth Annual Conference on Neural Information Processing Systems}, 2024. [Online]. Available: \url{https://openreview.net/forum?id=bZpZMdY1sj}
\BIBentrySTDinterwordspacing

\bibitem{khaliq2024vlad}
A.~Khaliq, M.~Xu, S.~Hausler, M.~Milford, and S.~Garg, ``Vlad-buff: burst-aware fast feature aggregation for visual place recognition,'' in \emph{European Conference on Computer Vision}.\hskip 1em plus 0.5em minus 0.4em\relax Springer, 2024, pp. 447--466.

\bibitem{ali2024boq}
A.~Ali-bey, B.~Chaib-draa, and P.~Gigu{\`e}re, ``Boq: A place is worth a bag of learnable queries,'' in \emph{Proceedings of the IEEE/CVF Conference on Computer Vision and Pattern Recognition}, 2024, pp. 17\,794--17\,803.

\bibitem{warburg2020mapillary}
F.~Warburg, S.~Hauberg, M.~Lopez-Antequera, P.~Gargallo, Y.~Kuang, and J.~Civera, ``Mapillary street-level sequences: A dataset for lifelong place recognition,'' in \emph{Proceedings of the IEEE/CVF conference on computer vision and pattern recognition}, 2020, pp. 2626--2635.

\bibitem{berton2022rethinking}
G.~Berton, C.~Masone, and B.~Caputo, ``Rethinking visual geo-localization for large-scale applications,'' in \emph{Proceedings of the IEEE/CVF Conference on Computer Vision and Pattern Recognition}, 2022, pp. 4878--4888.

\bibitem{ali2022gsv}
A.~Ali-bey, B.~Chaib-draa, and P.~Giguere, ``Gsv-cities: Toward appropriate supervised visual place recognition,'' \emph{Neurocomputing}, vol. 513, pp. 194--203, 2022.

\bibitem{alibeigi2023zenseact}
M.~Alibeigi, W.~Ljungbergh, A.~Tonderski, G.~Hess, A.~Lilja, C.~Lindstr{\"o}m, D.~Motorniuk, J.~Fu, J.~Widahl, and C.~Petersson, ``Zenseact open dataset: A large-scale and diverse multimodal dataset for autonomous driving,'' in \emph{Proceedings of the IEEE/CVF International Conference on Computer Vision}, 2023, pp. 20\,178--20\,188.

\bibitem{berton2025meshvpr}
G.~Berton, L.~Junglas, R.~Zaccone, T.~Pollok, B.~Caputo, and C.~Masone, ``Meshvpr: Citywide visual place recognition using 3d meshes,'' in \emph{European Conference on Computer Vision}.\hskip 1em plus 0.5em minus 0.4em\relax Springer, 2025, pp. 321--339.

\bibitem{berton2024earthloc}
G.~Berton, A.~Stoken, B.~Caputo, and C.~Masone, ``Earthloc: Astronaut photography localization by indexing earth from space,'' in \emph{Proceedings of the IEEE/CVF Conference on Computer Vision and Pattern Recognition}, 2024, pp. 12\,754--12\,764.

\bibitem{kolmet2022text2pos}
M.~Kolmet, Q.~Zhou, A.~O{\v{s}}ep, and L.~Leal-Taix{\'e}, ``Text2pos: Text-to-point-cloud cross-modal localization,'' in \emph{Proceedings of the IEEE/CVF Conference on Computer Vision and Pattern Recognition}, 2022, pp. 6687--6696.

\bibitem{keetha2023anyloc}
N.~Keetha, A.~Mishra, J.~Karhade, K.~M. Jatavallabhula, S.~Scherer, M.~Krishna, and S.~Garg, ``Anyloc: Towards universal visual place recognition,'' \emph{IEEE Robotics and Automation Letters}, 2023.

\bibitem{vivanco2024geoclip}
V.~Vivanco~Cepeda, G.~K. Nayak, and M.~Shah, ``Geoclip: Clip-inspired alignment between locations and images for effective worldwide geo-localization,'' \emph{Advances in Neural Information Processing Systems}, vol.~36, 2024.

\bibitem{xu2024addressclip}
S.~Xu, C.~Zhang, L.~Fan, G.~Meng, S.~Xiang, and J.~Ye, ``Addressclip: Empowering vision-language models for city-wide image address localization,'' in \emph{European Conference on Computer Vision}.\hskip 1em plus 0.5em minus 0.4em\relax Springer, 2024, pp. 76--92.

\bibitem{cattaneo2020global}
D.~Cattaneo, M.~Vaghi, S.~Fontana, A.~L. Ballardini, and D.~G. Sorrenti, ``Global visual localization in lidar-maps through shared 2d-3d embedding space,'' in \emph{2020 IEEE International Conference on Robotics and Automation (ICRA)}.\hskip 1em plus 0.5em minus 0.4em\relax IEEE, 2020, pp. 4365--4371.

\bibitem{ge2024bev2pr}
F.~Ge, Y.~Zhang, S.~Shen, Y.~Wang, W.~Hu, and J.~Gao, ``Bev2pr: Bev-enhanced visual place recognition with structural cues,'' \emph{arXiv preprint arXiv:2403.06600}, 2024.

\bibitem{leyva2023data}
M.~Leyva-Vallina, N.~Strisciuglio, and N.~Petkov, ``Data-efficient large scale place recognition with graded similarity supervision,'' in \emph{Proceedings of the IEEE/CVF Conference on Computer Vision and Pattern Recognition}, 2023, pp. 23\,487--23\,496.

\bibitem{he2016deep}
K.~He, X.~Zhang, S.~Ren, and J.~Sun, ``Deep residual learning for image recognition,'' in \emph{Proceedings of the IEEE conference on computer vision and pattern recognition}, 2016, pp. 770--778.

\bibitem{schroff2015facenet}
F.~Schroff, D.~Kalenichenko, and J.~Philbin, ``Facenet: A unified embedding for face recognition and clustering,'' in \emph{Proceedings of the IEEE conference on computer vision and pattern recognition}, 2015, pp. 815--823.

\bibitem{douze2024faiss}
M.~Douze, A.~Guzhva, C.~Deng, J.~Johnson, G.~Szilvasy, P.-E. Mazar{\'e}, M.~Lomeli, L.~Hosseini, and H.~J{\'e}gou, ``The faiss library,'' \emph{arXiv preprint arXiv:2401.08281}, 2024.

\bibitem{shao2023global}
S.~Shao, K.~Chen, A.~Karpur, Q.~Cui, A.~Araujo, and B.~Cao, ``Global features are all you need for image retrieval and reranking,'' in \emph{Proceedings of the IEEE/CVF International Conference on Computer Vision}, 2023, pp. 11\,036--11\,046.

\bibitem{sarlin2020superglue}
P.-E. Sarlin, D.~DeTone, T.~Malisiewicz, and A.~Rabinovich, ``Superglue: Learning feature matching with graph neural networks,'' in \emph{Proceedings of the IEEE/CVF conference on computer vision and pattern recognition}, 2020, pp. 4938--4947.

\bibitem{sun2021loftr}
J.~Sun, Z.~Shen, Y.~Wang, H.~Bao, and X.~Zhou, ``Loftr: Detector-free local feature matching with transformers,'' in \emph{Proceedings of the IEEE/CVF conference on computer vision and pattern recognition}, 2021, pp. 8922--8931.

\bibitem{lindenberger2023lightglue}
P.~Lindenberger, P.-E. Sarlin, and M.~Pollefeys, ``Lightglue: Local feature matching at light speed,'' in \emph{Proceedings of the IEEE/CVF International Conference on Computer Vision}, 2023, pp. 17\,627--17\,638.

\bibitem{wang2024efficient}
Y.~Wang, X.~He, S.~Peng, D.~Tan, and X.~Zhou, ``Efficient loftr: Semi-dense local feature matching with sparse-like speed,'' in \emph{Proceedings of the IEEE/CVF Conference on Computer Vision and Pattern Recognition}, 2024, pp. 21\,666--21\,675.

\bibitem{lee2022correlation}
S.~Lee, H.~Seong, S.~Lee, and E.~Kim, ``Correlation verification for image retrieval,'' in \emph{Proceedings of the IEEE/CVF conference on computer vision and pattern recognition}, 2022, pp. 5374--5384.

\bibitem{xue2022efficient}
F.~Xue, I.~Budvytis, D.~O. Reino, and R.~Cipolla, ``Efficient large-scale localization by global instance recognition,'' in \emph{Proceedings of the IEEE/CVF Conference on Computer Vision and Pattern Recognition}, 2022, pp. 17\,348--17\,357.

\bibitem{wang2022transvpr}
R.~Wang, Y.~Shen, W.~Zuo, S.~Zhou, and N.~Zheng, ``Transvpr: Transformer-based place recognition with multi-level attention aggregation,'' in \emph{Proceedings of the IEEE/CVF Conference on Computer Vision and Pattern Recognition}, 2022, pp. 13\,648--13\,657.

\bibitem{zhang2023etr}
H.~Zhang, X.~Chen, H.~Jing, Y.~Zheng, Y.~Wu, and C.~Jin, ``Etr: An efficient transformer for re-ranking in visual place recognition,'' in \emph{Proceedings of the IEEE/CVF Winter Conference on Applications of Computer Vision}, 2023, pp. 5665--5674.

\bibitem{zhu2023r2former}
S.~Zhu, L.~Yang, C.~Chen, M.~Shah, X.~Shen, and H.~Wang, ``R2former: Unified retrieval and reranking transformer for place recognition,'' in \emph{Proceedings of the IEEE/CVF Conference on Computer Vision and Pattern Recognition}, 2023, pp. 19\,370--19\,380.

\bibitem{hausler2021patch}
S.~Hausler, S.~Garg, M.~Xu, M.~Milford, and T.~Fischer, ``Patch-netvlad: Multi-scale fusion of locally-global descriptors for place recognition,'' in \emph{Proceedings of the IEEE/CVF conference on computer vision and pattern recognition}, 2021, pp. 14\,141--14\,152.

\bibitem{luo2023bevplace}
L.~Luo, S.~Zheng, Y.~Li, Y.~Fan, B.~Yu, S.-Y. Cao, J.~Li, and H.-L. Shen, ``Bevplace: Learning lidar-based place recognition using bird's eye view images,'' in \emph{Proceedings of the IEEE/CVF International Conference on Computer Vision}, 2023, pp. 8700--8709.

\bibitem{zaffar2024estimation}
M.~Zaffar, L.~Nan, and J.~F. Kooij, ``On the estimation of image-matching uncertainty in visual place recognition,'' in \emph{Proceedings of the IEEE/CVF Conference on Computer Vision and Pattern Recognition}, 2024, pp. 17\,743--17\,753.

\bibitem{li2024vxp}
Y.-J. Li, M.~Gladkova, Y.~Xia, R.~Wang, and D.~Cremers, ``Vxp: Voxel-cross-pixel large-scale image-lidar place recognition,'' \emph{arXiv preprint arXiv:2403.14594}, 2024.

\bibitem{yang2024depth}
L.~Yang, B.~Kang, Z.~Huang, X.~Xu, J.~Feng, and H.~Zhao, ``Depth anything: Unleashing the power of large-scale unlabeled data,'' in \emph{Proceedings of the IEEE/CVF Conference on Computer Vision and Pattern Recognition}, 2024, pp. 10\,371--10\,381.

\bibitem{geiger2012we}
A.~Geiger, P.~Lenz, and R.~Urtasun, ``Are we ready for autonomous driving? the kitti vision benchmark suite,'' in \emph{2012 IEEE conference on computer vision and pattern recognition}.\hskip 1em plus 0.5em minus 0.4em\relax IEEE, 2012, pp. 3354--3361.

\bibitem{zhang2024mvse}
J.~Zhang, Y.~Zhang, L.~Rong, R.~Tian, and S.~Wang, ``Mvse-net: A multi-view deep network with semantic embedding for lidar place recognition,'' \emph{IEEE Transactions on Intelligent Transportation Systems}, 2024.

\bibitem{lee2022patchworkpp}
S.~Lee, H.~Lim, and H.~Myung, ``{Patchwork++: Fast and robust ground segmentation solving partial under-segmentation using 3D point cloud},'' in \emph{Proc. IEEE/RSJ Int. Conf. Intell. Robots Syst.}, 2022, pp. 13\,276--13\,283.

\bibitem{bradski2008learning}
G.~Bradski and A.~Kaehler, \emph{Learning OpenCV: Computer vision with the OpenCV library}.\hskip 1em plus 0.5em minus 0.4em\relax " O'Reilly Media, Inc.", 2008.

\end{thebibliography}

\clearpage
\section*{Appendix}
\subsection{Data Preprocessing}
Data preprocessing is essential in cross-modal VPR, as it seeks to reduce modality differences and improve the overlap in visual content. Utilizing the methodologies outlined in ModaLink \cite{xie2024modalink} and Lip-Loc \cite{shubodh2024lip}, we perform a cropping of the RGB image along the horizontal direction at a predetermined position. In the KITTI dataset, an RGB image originally sized at (1226, 370) is cropped to (1226, 205). The complete 360° point cloud is cropped according to the camera's horizontal field-of-view, ensuring optimal visual content alignment between the processed images and point clouds. This step minimizes content discrepancies in retrieval processes.

Subsequently, range images are produced from point clouds utilizing established techniques, thereby accurately representing point cloud information in the vertical direction. The range images are subsequently aligned with the RGB images. Unobstructed information is integrated into both modalities by generating BEV representations. The cropped RGB image undergoes processing via a monocular depth prediction model to generate a depth prediction map. The Sobel operator is utilized for edge detection on the depth map to filter out ambiguous depth information along the edges. The refined depth map is subsequently transformed into a pseudo point cloud in the LiDAR coordinate system through the application of camera intrinsics, camera extrinsics, and the extrinsic parameters of LiDAR.

\begin{figure*}[t]
    \centering
    \includegraphics[scale=0.32]{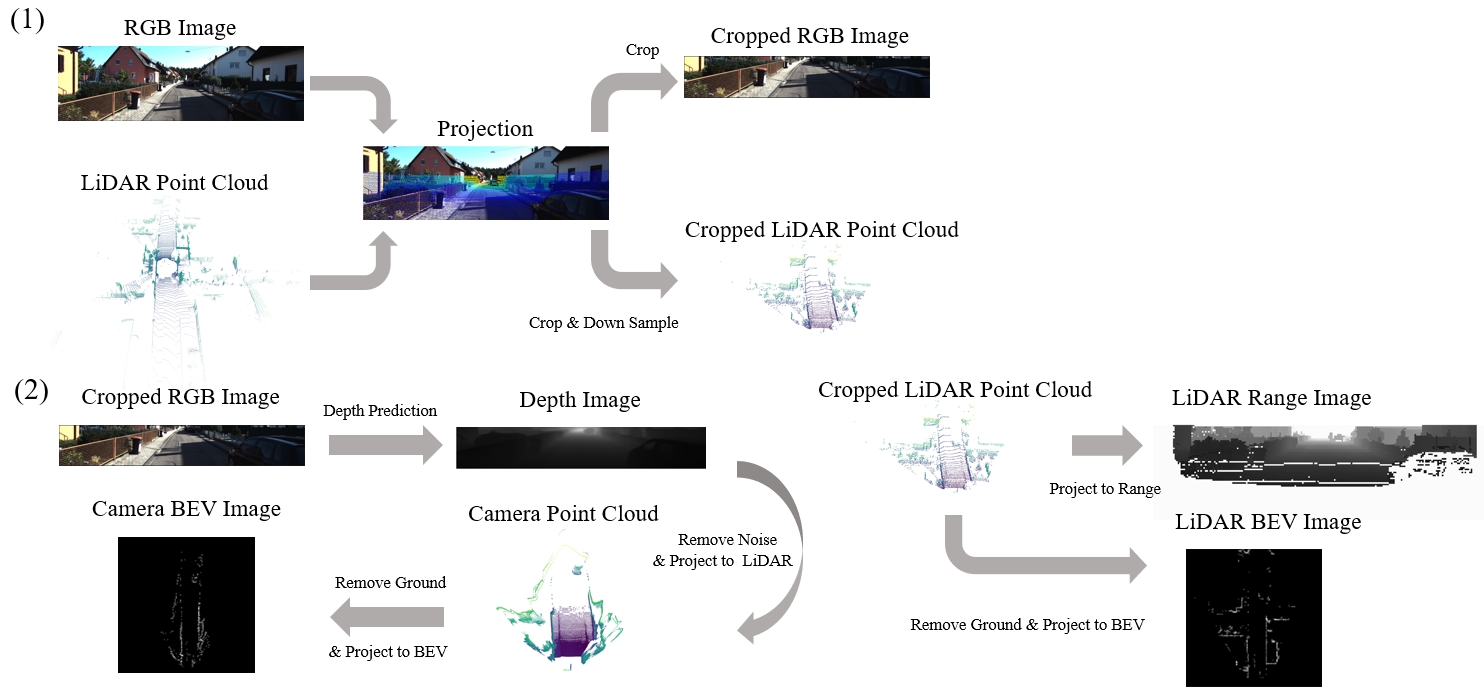}
    \caption{The data preprocessing pipeline. (1) Top is the cropping process (2) Bottom displays the BEV image generation process.}
    \label{fig:datapreprocessing}
\end{figure*}

After acquiring the camera point clouds and LiDAR point clouds, ground points are eliminated to reduce noise. The point clouds are subsequently converted into BEV representations from a top-down perspective. In the I2P-Rec's setup \cite{zheng2023i2p}, the LiDAR coordinate system functions as the reference framework, with point clouds confined to defined coordinate ranges: the x-axis spans [0, 51.2m], the y-axis ranges from -25.6m to 25.6m, and the z-axis extends from -5.0m to 5.0m. The voxel size for the BEV image is established at 0.4m, yielding a final BEV image resolution of [128, 128]. This BEV representation effectively conveys orientation and scene topology information, clearly illustrating the spatial distribution of buildings and streets in autonomous driving contexts. The data preprocessing pipeline is displayed on the Fig. \ref{fig:datapreprocessing}.

\subsection{Additional Ablation}
Our ablation studies primarily focus on evaluating the effects of several key factors: the base margin of the generalized triplet loss, the threshold for distance similarity in the points average distance method, and the impact of ground removal and noise removal algorithms. Both quantitative results and visualizations are provided to support our analysis.

\begin{table}[!t]
    \centering
    \caption{ABLATION STUDIES ON BASE MARGIN}
    \label{tab:AblationBM}
    \resizebox{6.5cm}{!}{
    \begin{tabular}{c|c|c}
    \hline
        & KITTI-00 & KITTI-02 \\ \hline
        Base Margin & R@1 / R@5 / R@1\% & R@1 / R@5 / R@1\% \\ \hline
        0.4 & 99.10 / 99.93 / 100.0 & 94.79 / 98.03 / 99.61 \\ 
        0.5 & 99.05 / 99.87 / 100.0  & 95.04 / 98.05 / 99.85 \\ 
        0.6 & 99.03 / 99.91 / 100.0 & 95.07 / 98.22 / 99.66 \\ 
        0.7 & 98.77 / 99.82 / 100.0 & 94.70 / 97.88 / 99.68 \\ 
        0.8 & 99.19 / 99.91 / 100.0 & 95.73 / 98.46 / 99.72 \\ \hline
    \end{tabular}}
\end{table}

As shown in Table \ref{tab:AblationBM}, the performance remains consistent despite changes in the base margin, indicating that the absolute scale of the delta value between the relative sample and the anchor sample is not critical—rather, it is the delta value itself that plays a crucial role.

\begin{table}[!t]
    \centering
    \caption{ABLATION STUDIES ON DISTANCE THRESHOLD}
    \label{tab:AblationDT}
    \resizebox{6.5cm}{!}{
    \begin{tabular}{c|c|c}
    \hline
        & KITTI-00 & KITTI-02 \\ \hline
        Dist\_sim\_th(m) & R@1 / R@5 / R@1\% & R@1 / R@5 / R@1\% \\ \hline
        2.5 & 98.00 / 99.96 / 100.0 & 92.86 / 97.66 / 99.61 \\ 
        5 & 98.61 / 99.93 / 100.0 & 95.19 / 98.13 / 99.76  \\ 
        7.5 & 99.03 / 99.91 / 100.0 & 95.07 / 98.22 / 99.66 \\ 
        10 & 98.68 / 99.80 / 100.0 & 94.79 / 97.77 / 99.61 \\ \hline
    \end{tabular}}
\end{table}

Additionally, Table \ref{tab:AblationDT} demonstrates that the optimal performance is achieved when the threshold for distance similarity is set to 7.5m. When this value is too high, it includes samples without any visual content overlap, leading to false positives. Conversely, when the value is too low, it disregards meaningful relationships between samples with actual appearance similarities.

\begin{figure*}[!t]
    \centering
    \includegraphics[scale=0.3]{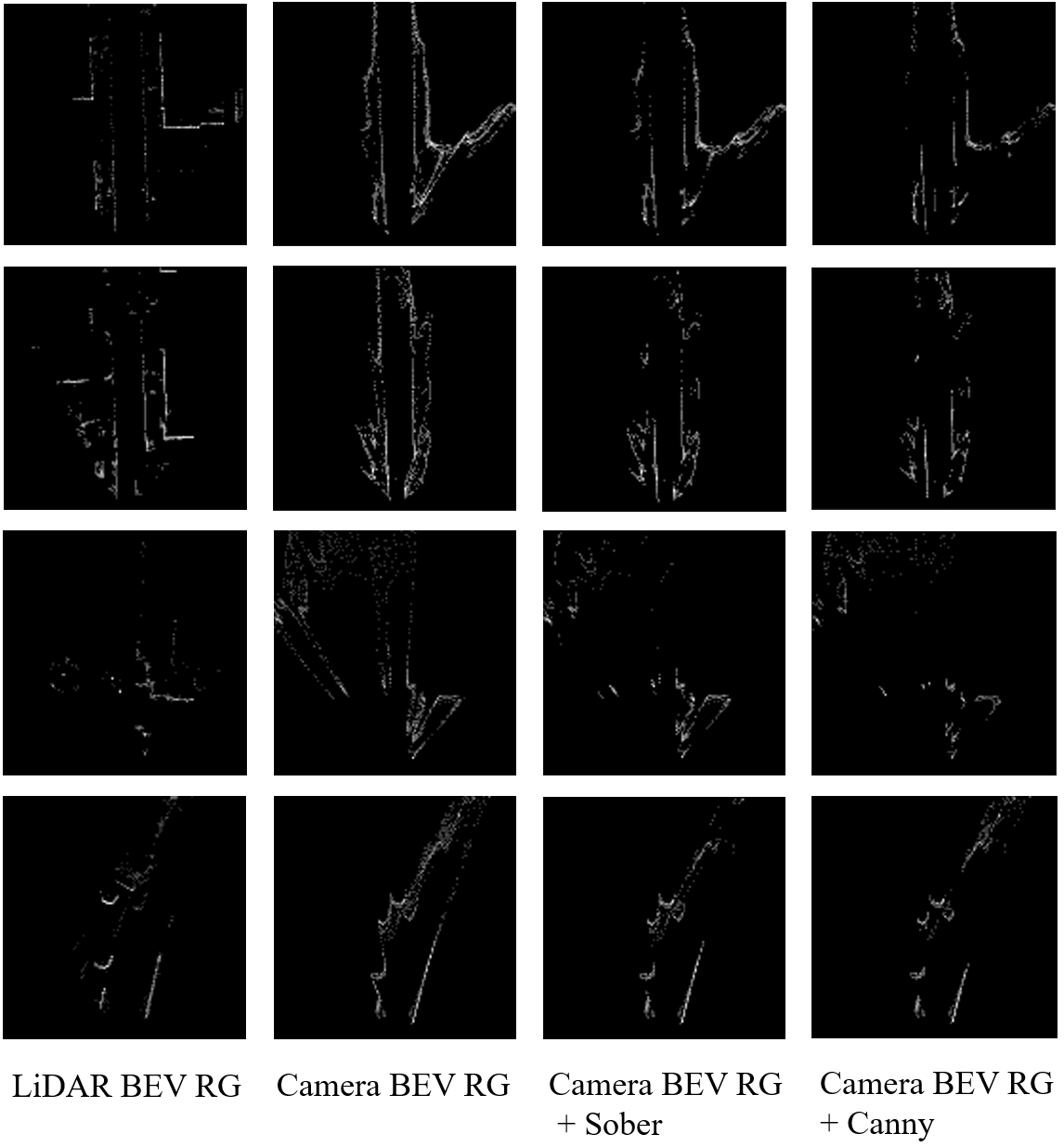}
    \caption{Examples of LiDAR BEV and Camera BEV with different processing.}
    \label{fig:NoiseRemove}
\end{figure*}

\begin{table}[!t]
    \centering
    \caption{ABLATION STUDIES ON GROUND-REMOVAL (RG) AND NOISE-REMOVAL (RN)}
    \label{tab:AblationR}
    \resizebox{7.5cm}{!}{
    \begin{tabular}{c|c|c}
    \hline
        & KITTI-00 & KITTI-02 \\ \hline
        Method & R@1 / R@5 / R@1\% & R@1 / R@5 / R@1\% \\ \hline
        Vanilla & 97.82 / 99.82 / 100.0 & 89.62 / 96.14 / 99.68 \\
        RG & 97.73 / 99.65 / 100.00 & 93.03 / 97.34 / 99.53 \\
        RG + RN (Sober) & 99.03 / 99.91 / 100.00 & 95.07 / 98.22 / 99.66 \\
        RG + RN (Canny) & 98.61 / 99.78 / 100.00 & 94.92 / 98.13 / 99.76 \\ \hline
    \end{tabular}}
\end{table}

At last, we emphasize the importance of the ground and noise removal process. To remove ground points from the point cloud, we employ the Patchwork++ \cite{lee2022patchworkpp} algorithm, while noise in the Camera BEV image, caused by ambiguous depth predictions at object edges, is detected using the Sober calculator provided by OpenCV \cite{bradski2008learning}. This method approximates the noise area and marks it as invalid during back-projection. The Canny edge detection algorithm is also employed to detect object edges and expand them based on depth values. In Table \ref{tab:AblationR}, we observe that removing both the ground points and noise simultaneously significantly improves the re-rank method’s performance, with the Sober calculator yielding the best results. Finally, Fig. \ref{fig:NoiseRemove} demonstrates that after removing the ground, the LiDAR and Camera BEV images become more similar, and using the Sober calculator effectively removes noise while preserving essential visual cues, outperforming the Canny edge detection method.
\end{document}